\documentclass{article}

\usepackage{microtype}
\usepackage{graphicx}
\usepackage{subfigure}
\usepackage{booktabs} 
\usepackage[utf8]{inputenc} 
\usepackage[T1]{fontenc}    
\usepackage{url}            
\usepackage{booktabs}       
\usepackage{amsfonts}       
\usepackage{nicefrac}       
\usepackage{microtype}      
\usepackage{url}
\usepackage{natbib}
\usepackage{graphicx}
\usepackage{amsmath,amssymb,amsthm}
\usepackage{mkolar_definitions}
\usepackage{natbib}
\usepackage{multirow}
\usepackage{algorithm}
\usepackage{algorithmic}
\usepackage{subfigure}
\usepackage{wrapfig}

\DeclareMathOperator*{\sign}{\textit{sign}}

\newcommand{\bbe}{\mathbb E}
\newcommand{\wmin}{w_{\star}}

\newcommand{\Ft}{\nabla F_t}
\newcommand{\gti}{\tilde g}
\usepackage{hyperref}

\usepackage[accepted]{icml_arxiv}

\icmltitlerunning{Trajectory Normalized Gradients for Distributed Optimization}

\begin{document}

\twocolumn[
\icmltitle{Trajectory Normalized Gradients for Distributed Optimization}




\begin{icmlauthorlist}
\icmlauthor{Jianqiao Wangni}{to}
\icmlauthor{Ke Li}{ucb}
\icmlauthor{Jianbo Shi}{to}
\icmlauthor{Jitendra Malik}{ucb}


\end{icmlauthorlist}

\icmlaffiliation{to}{University of Pennsylvania}
\icmlaffiliation{ucb}{University of California, Berkeley}

\icmlcorrespondingauthor{Jianqiao  Wangni}{wnjq@seas.upenn.edu}

\icmlkeywords{Machine Learning, ICML}

\vskip 0.3in
]



\printAffiliationsAndNotice{}  

\begin{abstract}
Recently, researchers proposed various low-precision gradient compression, for efficient communication in large-scale distributed optimization. Based on these work, we try to reduce the communication complexity from a new direction. We pursue an ideal bijective mapping between two spaces of gradient distribution, so that the mapped gradient carries greater information entropy after the compression. In our setting, all servers should share a reference gradient in advance, and they communicate via the normalized gradients, which are the subtraction or quotient, between current gradients and the reference. To obtain a reference vector that yields a stronger signal-to-noise ratio, dynamically in each iteration, we extract and fuse information from the past trajectory in hindsight, and search for an optimal reference for compression. We name this to be the trajectory-based normalized gradients (TNG). It bridges the research from different societies, like coding, optimization, systems, and learning. It is easy to implement and can universally combine with existing algorithms. Our experiments on benchmarking hard non-convex functions, convex problems like logistic regression demonstrate that TNG is more compression-efficient for communication of distributed optimization of general functions.

\end{abstract}

\section{Introduction}

For large-scale machine learning, the distributed optimization algorithm makes a significant contribution to improving the performance and scalability \cite{bottou2018optimization}.  An almost necessary technique to process massive amounts of data in parallel is to divide data to different servers within a computation cluster, and these servers will provide local gradients and perform model synchronization using various communication protocols. In a centralized and synchronous setting, that all servers transmit their local gradients back to the main server, then the main server computes an updated parameter value and broadcast to others. As the system grows in size, the synchronization procedure is likely to be slowed by the network capacity and latency. 
A popular way to reduce the communication cost is to transmit compressed gradients, i.e. the low-bit representation. There are studies using low-bit quantization \cite{alistarh2016qsgd}, ternary representation in $\{-1,0,1\}$ \cite{zhou2016dorefa, wen2017terngrad},  sparsified vectors  \cite{wang2018atomo, alistarh2018convergence, wangni2018gradient}, top-K important coordinates \cite{aji2017sparse} or even only using signs of gradients \cite{bernstein2018signsgd}. 
Plus, the compression error in previous iterations can be accumulated as \cite{wu2018error, stich2018sparsified} to compensate the gradients. 

The compression error is still far from rigorously studied. Most of the above works focused on the stochastic gradient descent (SGD) \cite{zhang2004solving, bottou2010large} and for training deep neural networks, the objective function of which are naturally robust to optimization with noisy gradients \cite{jin2017escape,kleinberg2018alternative}. However, for wider range problems, optimization for convex and strongly-convex problems are sensitive to gradient noise, which partially explains that variance-reduced SGD \cite{johnson2013accelerating} and quasi-Newton methods \cite{wright1999numerical} strongly outperform vanilla SGD. Under these settings, the convergence rate will probably slow down linearly by the compression, then there are theoretically no savings in terms of communication cost. Therefore, it is imperative to characterize the compression error more in-depth.

A natural motivation is that the compression error strongly depends on the gradient distribution, in addition to the compression algorithm itself. For example, the Huffman coding favors the distribution of literals occurrence being skewed \cite{cormen2009introduction}; frequency-domain image codings are effective since the low-frequency and high-frequency parts have an unbalanced distribution of sensitivity  to human eyes \cite{szeliski2010computer}. Perhaps, just like the \textit{no free lunch theorem}, there will be no effective compression without further distribution properties to apply.

Motivated by this, we propose to effectively adjust or normalize the gradient distribution before compressing them; ideally, having a distribution of standard Gaussian. The problem is different from a conventional sense of communication, from several perspectives: 1) the ultimate target of these rounds of gradient exchange is to improve the optimization in outer framework, 2), the information is generated by the optimization algorithm, which can be modified to adapt the encoding and decoding, 3) the past gradients shared in advance may be used to accelerate, like quasi-Newton algorithms \cite{wright1999numerical} and Nesterov's momentum \cite{nesterov2013introductory}, and they naturally cost no extra communication.

The paper is arranged as follows: we will introduce the background and notations, then the motivation of normalized gradients; we give some implementation options on the idea and we evaluate the idea on different problems.


\section{Background}

Denote $F(w)$ as the objective function, and $w$ is the parameter to be optimized. For convenience, we assume that the objective has a finite average formulation over $N$ data points, and each loss function is denoted as $f_n(w)$. In the $t^{\text{th}}$ round, a descent vector $g(w_t)$ based on the current parameter $w_t$
\begin{align*}
\min_w \quad F(w) := \frac{1}{N} \sum_{n=1}^N f_n (w), \quad w \in \mathbb R^D
\end{align*}
The descent vector $g(w_t)$ has to an unbiased estimation of the gradient $\nabla F(w)$, and has a bounded variance term to assure the convergence. 
A typical strategy for stochastic gradient descent (SGD), is to have an index  $n_t \in [N]$ uniformly sampled from the data set and take a step as 
\begin{align*}
w_{t+1} =w_{t} - \eta_t g(w_{t}),\quad g_t(w_t) = \nabla f_{n_t} (w_t), 
\end{align*}
here $\eta_t$ is the step-size for this iteration. 

In a distributed computation model, where we assume $M$ servers are available for the optimization task. Let each server has its own share of the whole training dataset, say server $m$ has $\Omega_m \subseteq [N]$, 
and they provide an unbiased estimation of gradients by averaging together,
\begin{align*}
 g(w)= \frac{1}{M} \sum_m^M g^m(w), \quad \bbe [g^m(w)]=\frac{1}{|\Omega_m|} \sum_{i \in \Omega_m} \nabla f_n(w).
\end{align*}

In each around, server $m$ calculates its unbiased estimation of the gradient $g^m(w_t) =\nabla f_{i_{m,t}} (w) $ where $i_{m,t}$ is randomly sampled, based on partial data from its memory, then transmits the gradients to main server for synchronization, during which the main server average over all gradients and updates the parameter $w_{t+1} =w_{t} - \eta_t g(w_{t})$, 
and broadcasts it back to all servers. 

Previous research on compressed gradient assumes that there exists a coding strategy $Q: \mathbb R^{D} \rightarrow  \Psi^{D}$ to compress the gradient vector, where $\Psi$ is the available set for representing a number in $\mathbb R$. Then each server only needs to update its gradient using a compressed vector, and the overall algorithm behaves like
\begin{align}\label{eq:qsgd}
w_{t+1} =w_{t} - \eta_t \frac{1}{M}\sum_m  Q[g^m(w_t)].
\end{align}
Besides, an ideal  design of compression should be unbiased, so that
\begin{align*}
\bbe Q[g(w)]=\bbe [g(w)] = \nabla F(w).
\end{align*}

\subsection{Motivation}

Suppose we use an algorithm in Eq.(\ref{eq:qsgd}) and target for an $L$-smooth loss function $F$, and we assume that the compression error is random and independent of $g$.   The convergence rates for the methods designed above are strongly related to the optimization algorithm, especially the strategies of generating gradients in each iteration, as well as the assumptions (i.e. smoothness, Lipchitz continuity, convexity).   For convenience, we denote $g_t =g(w_t)$ and $\Ft= \nabla F(w_t)$.

\begin{assumption}\label{smooth}
We suppose that the loss function $F$ is differentiable and  $L$-smooth and  $\lambda$-strongly convex
\begin{align*}
&F(y)-F(x)- \nabla F(y)^{\top}(y-x) \leq \frac{L}{2} ||y-x||^2,\\
&F(y)-F(x)- \nabla F(x)^{\top}(y-x) \geq \frac{\lambda}{2} ||y-x||^2.
\end{align*}
\end{assumption}


We start with a simple inequality: based on the iteration $t$, the expected loss for the next iteration is bounded by
\begin{align*}
&\mathbb E \; [F(w_{t+1})]-F(w_t)\\
\leq  &\mathbb E\left[  \Ft^\top (w_{t+1}-w_t) + \frac{L}{2} \|w_{t+1}-w_t\|^2\right] \\
\leq & - \eta_t \| \Ft \|^2+\frac{L}{2} \eta_t^2 \bbe [\| g_t \|^2] + \frac{L}{2} \eta_t^2 \mathbb E   [\|Q[g_t]-g_t\|^2] .
\end{align*}
where we applied the smoothness property in the first inequality, and decomposed the variance in the second inequality. An optimal compression $Q$ is supposed to reduce the  variance from compression error in $Q[g_t]- g$. 

Although seldom studied in this area of communication-efficient distributed optimization, we notice that the compression error is largely affected by the gradient distribution.  Different compression strategies favor different kinds of distribution, whether it is long-tail, or strongly-concentrated like sub-Gaussian, or weakly-concentrated like sub-exponential.  For example, gradient quantization approaches \cite{alistarh2016qsgd} favors gradients with uniformly distributed elements within the quantization range; but differently, if one uses the gradient sparsification technique \cite{wangni2018gradient} as the compression $Q$, then reversely, a strong skewness of gradients implies that the communication could be saved more. 

\section{Normalized Gradients}

We try to address the problem by adjusting or normalizing the gradient distribution by past trajectories, since they have been transmitted so do not incur additional communication cost in this round. We refere to the adjusted gradient to be Trajectory Normalized Gradient (TNG). The communication protocol can be generally described as: we wish to let all servers share a gradient vector $\gti$ that approximate $g_t$ in advance. For sending the gradients, each server transmits the normalized gradients, i.e. the difference between $g_t - \gti$. 
Each server could send gradients using compressed TNG $s(w_t)$; then upon receiving $r(w_t)$, a server uses the following procedure to decode the gradient $v(w_t)$ as
\begin{align}
 r(w_t)=Q [g_t - \gti], \quad v(w_t)=\gti+   r(w_t).
\end{align}
A simple understanding of $g_t - \gti$ is to view it as a zero-centered random variable, if $\gti =\Ft$, or a polynomial of the high order derivative,  if $\gti=g_{t-1}$, and the range for the normalized gradients is tighter by higher-order continuity. 
The distribution of $g_t - \gti$ and $g$ depends on the model, data and the optimization algorithm itself. If they follow the same distribution, only different in magnitude by a factor of  $C_{nz}\ll 1 $, clearly, the compression on $g_t - \gti$ yields a smaller error. 
By taking logarithms of gradients vectors $\gti$ and $g_t$ before performing the coding above, we get a form that 
\begin{align}
 r(w_t)=Q [g_t ./\gti],  \quad v(w_t)=\gti \odot   r(w_t),
\end{align}
where $\odot$ is the element-wise product and $./$ takes the element-wise quotient. If these two procedures are combined, we get a normalization form of
\begin{align*}
 r(w_t)=Q [(g_t- \gti) ./\gti'],  \quad v(w_t)=\gti' \odot   r(w_t)+ \gti,
\end{align*}
where $\gti'$ is a second reference vector. 
We also not that $\gti$ could be shared through a round of broadcast, from the main server, it could also be explicitly shared, for example, using a predefined protocol to update $\gti$ from the gradient vectors that these servers received from previous iterations.

\subsection{Reference Vectors}
The key requirement is choose $\gti$ appropriately so that $g_t - \gti$ follows a normalized distribution than $g_t$ for less compression error from $Q$. General normalization request the mean vector to be pre-known, which actually cause much trouble, as in each iteration of $g_t$ being updated, it has different means $\Ft$. The calculation of $\Ft$ can be assumed to be basically impossible, as it takes much more computation (linear to data numbers) than calculating gradient from a mini-batch as SGD. Here we reach an interesting problem about how to approximate the mean of stochastic gradient to make it actually normalized.

A simple approach is to take $\gti= \textit{mean}(g)\textit{ones}(D)$ where $\textit{mean}(g)$ is the average value of all elements in $g$. This will reduce the variance of $g$ from an inequality
\begin{align}
\bbe ||a - \bbe[a]||^2 \leq \bbe ||a ||^2,
\end{align}
for any random variable $a$. The only additional cost  is to transmit a single scalar $\textit{mean}(g)$, which is ignorable compared to transmitting a $D$-dimensional vector. 

The formulation for $\gti$ can be inspired from other areas. For example, the stochastic variance-reduced gradient (SVRG) algorithm \cite{johnson2013accelerating}  gives a better estimation of gradients converges linearly converges on strongly-convex and smooth loss functions, where
\begin{align*}
w_{t+1} =w_{t} - \eta_t (\nabla f_{n_t}(w_{t}) - \nabla f_{n_t}(\tilde w)+ \nabla F(\tilde{w}))
\end{align*} 
where $\tilde{w}$ is a reference parameter which is generally chosen from a previous iteration. The full gradient $\nabla F(\tilde{w}))$, although cost much more compared to stochastic gradients, are not frequently updated. Once the full gradient is evaluated, it only costs one round of communication for many rounds of SGD steps. Based on the same intuition, the stochast averaging gradient \cite{schmidt2017minimizing} could be applied here. The difference is that, the main server can average gradients from all servers, and the gradients might be the compressed ones $v(w_t)$ from past iterations.

In another area of distributed optimization, the delay-tolerant optimization algorithm \cite{agarwal2011distributed} performs the following updates
\begin{align*}
w_{t+1} =w_{t} - \eta_t g(w_{t-\tau}),\quad  \tau \in [0,1,\cdots,\tau_{\text{max}}]
\end{align*}
As long as the staleness of the parameter, $\tau$ or $||w_{t-\tau} - w_t||^2$, is bounded.  The gradient as above can be the reference gradient $\gti$ since it is a close approximation to the current gradient. 

The fourth option is to use a two-stage compression strategy that, in each stage, the algorithm generates a compensate vector $Q^2(v)$ with shared vector $\gti$ to complement the first stage $Q^1(v)$ and $\gti^1$. To list all of them here:
\begin{align*}
\gti =\begin{cases}
&  \nabla f_{n_t}(w_{t}) - \nabla f_{n_t}(\tilde w)+ \nabla F(\tilde{w}) \\
& \Sigma_{\tau}^{\tau_{\text{max}}} v(w_{t-\tau})/ \tau_{\text{max}}\\
& \nabla g(w_{t-\tau}),\quad \tau \in [0,1,\cdots,\tau_{\text{max}}] \\
& \textit{mean}(g_t) \textit{ones}(D) \\
& \textit{mean}(g_t - Q^1(g_t- \gti^1)- \gti^1 ) \textit{ones}(D).
\end{cases}
\end{align*}
The reference vector can be updated frequently or occasionally depending on the easiness of visiting it, e.g. setting an update frequency of $\tau_{\text{max}}$ like the staleness synchronous protocol (SSP) \cite{ho2013more}.

\subsection{Gradient Compression}
There are many protocols available for compressing the normalized gradient $v=g^m_t- \gti$, as the literature introduced above. Here we take a strong compression coding strategy for an example, e.g. using the sign of each element \cite{wen2017terngrad, bernstein2018signsgd}.  
For communication, each server transmits a constant $R_t^m$ as the largest element of $v_t^m=g^m_t- \gti$, and each compressed element $Q[v_d]=\sign[v_d]$ derived from the $d^{\text{th}}$ element of $v$. 
For simplicity, we will often omit the subscripts $m$ and $t$.  
The magnitude information is encoded by the randomization process, and the unbiasedness of the compressed gradient would won't change $\bbe [Q[v]]=v$ in expectation. 
\begin{align*}
Q[v] =  R \sign[v] \odot z(v),\quad R=\max_{d} |v_d|.
\end{align*}
Denote $R$ as the largest element of $v$, 
and a binary vector $z (v) \in \{0,1\}^D$,  to indicate whether each element of $v=g^m_t- \gti$  to be compressed by its sign or simply zero.
\begin{align*}
\begin{cases}
&  P(z(v_d)=1)= \frac{|v_d|}{R} \\
& P(z(v_d)=1)= 1-\frac{|v_d|}{R},\quad \forall d \in [D] .
\end{cases}
\end{align*}
An example of compressed TNG is in Algorithm \ref{alg:psgd}.
In the following, we will characterize the optimality of the coding strategy above, that the probability vector should be proportional to magnitudes.
\begin{proposition}\label{coding_opt}
For $L$-smooth loss functions $F(w)$, setting $P(z(v_d)=\textit{sign}[v_d]) \propto |v_d| $ proposed above is the optimal sampling probability for ternary coding of $g_t- \gti$ in $\{-1,0,1\}$ for optimizing $\bbe [F(w_{t+1})]$.
\end{proposition}

\begin{algorithm*}[h] 
\caption{Trajectory Normalized Gradients via Ternary Coding and Delayed Gradients}
\label{alg:psgd}
\begin{algorithmic} [1]
\STATE Initialize the clock $t=0$ and initialize the weight $w_0$, and set $\gti=zeros(D)$.
\REPEAT
\STATE Each server $m$ calculates local gradient $g^m_t$, $R_m = \max_d |g^m_t -\gti|_d$ and the vector $p= |g^m_t|/R_m$.
\STATE Randomly sample a binary vector $z_t$ that $P(z(v_t)_d=1)=p_d$ and $P(z(v_t)_d=0)=1-p_d$.
\STATE Transmit the compresed gradients  $Q(g^m_t)= \sign(g^m_t)\odot z_t$ and $R^m_t$.
\STATE The main server average over the received gradients  $v_t=\frac{1}{M} \sum_{m=1}^M R^m_t Q(g^m_t)$ and broadcast.
\STATE Update the reference vector $\gti$, through main server broadcasting.
\UNTIL{convergence or the number of iteration reaches the maximum setting.}
\end{algorithmic}
\end{algorithm*}

\subsection{Convergence Analysis}

We do not focus on the specific constant of the convergence rate, since it depends on other factors and it is hard to provide a unified theorem that is both informative and tight. Here, we give a simple analysis of how the compression error affects the convergence rate.


\newcommand{\cond}{C}

\begin{lemma}\label{variance_bound}
We have the following assumption for the variance of stochastic gradient $g(\wmin)$ evaluated at the optimal point $\wmin$, $\bbe ||g(w_{\star})||^2 \leq \sigma$. Then
for loss functions that satisfy assumption \ref{smooth}, the variance of $g$ is bounded by 
\begin{align*}
\bbe ||g(w)||^2 \leq 4 L(F(w) - F(w_{\star})) + 2\sigma^2,
\end{align*}
\end{lemma}

This lemma gives a better bound on the gradient variance rather than directly assigning an upper bound to the variance, as it decreases as the optimization is going on. 
\begin{proposition}\label{as:decre_var}
For compressed normalized gradient in Algorithm \ref{alg:psgd}, we assume that there exists a constant $ C_{nz} \in (0,1]$ that for stochastic gradients $g_t$ on all servers, 
\begin{align*}
\bbe \| g_t-\gti]\|^2 \leq  C_{nz} \bbe \| g_t \|^2.
\end{align*}
\end{proposition}
We could always assure the proposition above to be satisfied. For example, 
we can set $ C_{nz}=1$ and $\gti= \textit{zeros}(D)$ and get $C_{nz}=1$, although it degenerates to a trivial case. For real applications, this assumption can be much better satisfied, since we have a large pool of available reference vectors that can be shared in so many ways, e.g. using reference vectors from in hindsight. As long as there is a need for trading computation for communication, this constant $ C_{nz}$ can be searched. The additional communication cost for this is to indicate which $\gti$ is used for this iteration. 
\begin{assumption}\label{as:quant_var}
We assume that the coding strategy has bounded compression error for $g_t -\gti$, that
\begin{align*}
\bbe \|Q [g_t-\gti]]\|^2 \leq  C_{nz} \bbe \| [g_t-\gti] \|^2.
\end{align*}
\end{assumption}
We denote $C_{q,nz}=C_q  C_{nz} +1$ as a compression constant for TNG, and implies  neccessary bits for communication.
\begin{lemma}\label{h_g}
The variance of $v(w_t)$ is bounded as, 
\begin{align*}
\bbe[||v(w_t) ||^2] \leq C_{q,nz}(2L ||F(w_t) - F(\wmin)||^2+ \sigma^2).
\end{align*}
\end{lemma}


\textbf{Remark: } 
We apply an inequality for two variables $x,y$, $\bbe ||x+y||^2 \leq 2 \bbe ||x||^2 +2 \bbe ||y||^2$, 
and decompose the variance using Assumption on compression error into
\begin{align*}
&\bbe[||v(w_t) ||^2] = \bbe[|| \gti+  [Q[ g_t - \gti]]- g_t +g_t ||^2] \\
&\leq 2\bbe[||\gti+  [Q[ g_t - \gti]]- g_t ||^2 + ||g_t ||^2] \\
&= \bbe[2C_q|| g_t - \gti ||^2+ 2||g_t ||^2] 
\end{align*}
 After applying the assumption about shrinkage of variance for normalization, we have the lemma.
%


\begin{theorem}\label{main}
For loss functions and TNG algorithms that satisfy \ref{smooth}; \ref{as:decre_var}, after enough iterations $t \geq t_0$, and the step size satisfies
\begin{align*}
&\eta_t = \frac{\alpha}{\lambda (t+\alpha \kappa)} \leq \frac{1}{2L},\quad \kappa=\frac{2  L C_{q,nz}}{\lambda},\\
&t_0=\frac{4 L C_{q,nz}}{\lambda} \left( \max \left(\frac{C_{q,nz}}{\lambda \sigma} ||w_0 - \wmin||^2,1 \right)-1 \right),
\end{align*}
here $\alpha$ is a constant and $\kappa$ behaves like the condition number, then the suboptimality is guaranteed as
\begin{align*}
\bbe ||w_{t+1} - \wmin||^2 =O \left(\frac{4 \alpha^2 \sigma^2}{\lambda^2} \frac{1}{t- t_0 +\alpha \kappa}\right).
\end{align*}
\end{theorem}
This is an adaptation of a general analysis of strongly-convex optimization \cite{nguyen2018sgd} to include compression error, and gives us a basic intuition about the factors of compression error affecting the convergence rate.

\newcommand{\xyopt}{(x_{\star},y_{\star})}
\newcommand{\csk}{C_{sk}}
\newcommand{\cskk}{C_{th}}

\begin{figure}[!htbp]
	\vspace{-10pt}
	\centering
	\subfigure [Ackley Function]
	{\includegraphics[width=0.5\textwidth]{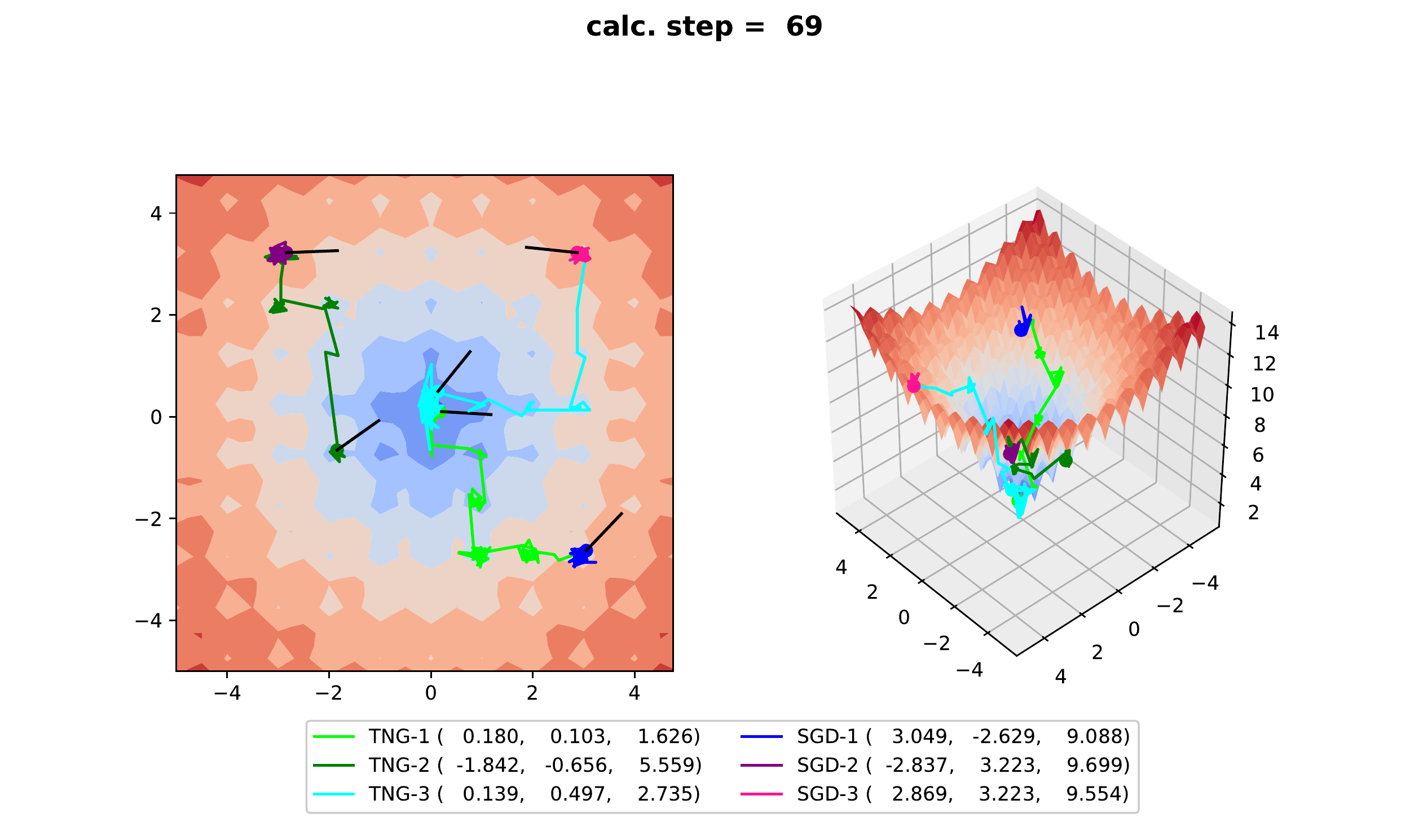}}
	%
	\vspace{-5pt}
	\subfigure [Booth Function]
	{\includegraphics[width=0.5\textwidth]{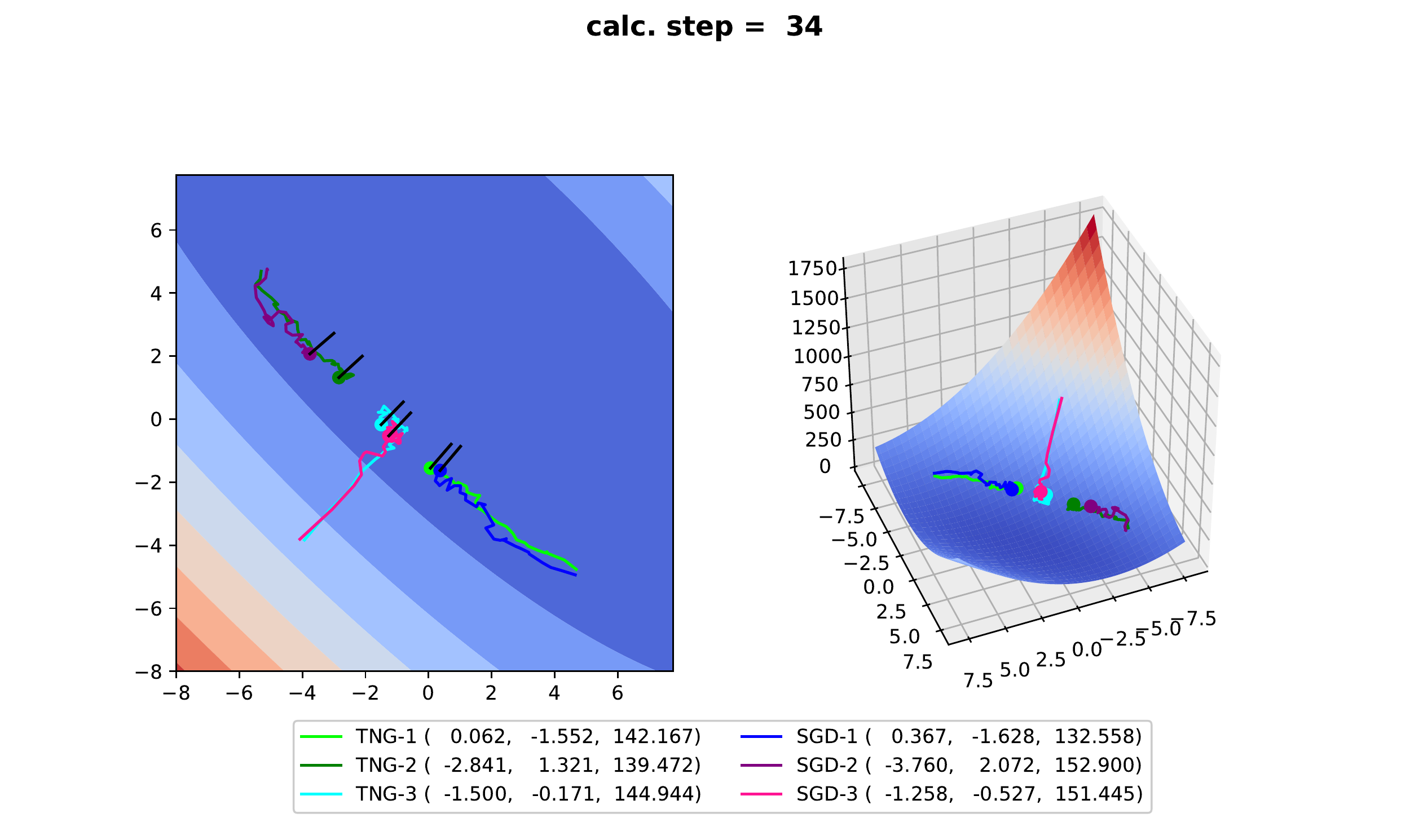}}
	%
	%
	\vspace{-5pt}
	\subfigure [Rosenbrock Function. ]
	{\includegraphics[width=0.5\textwidth]{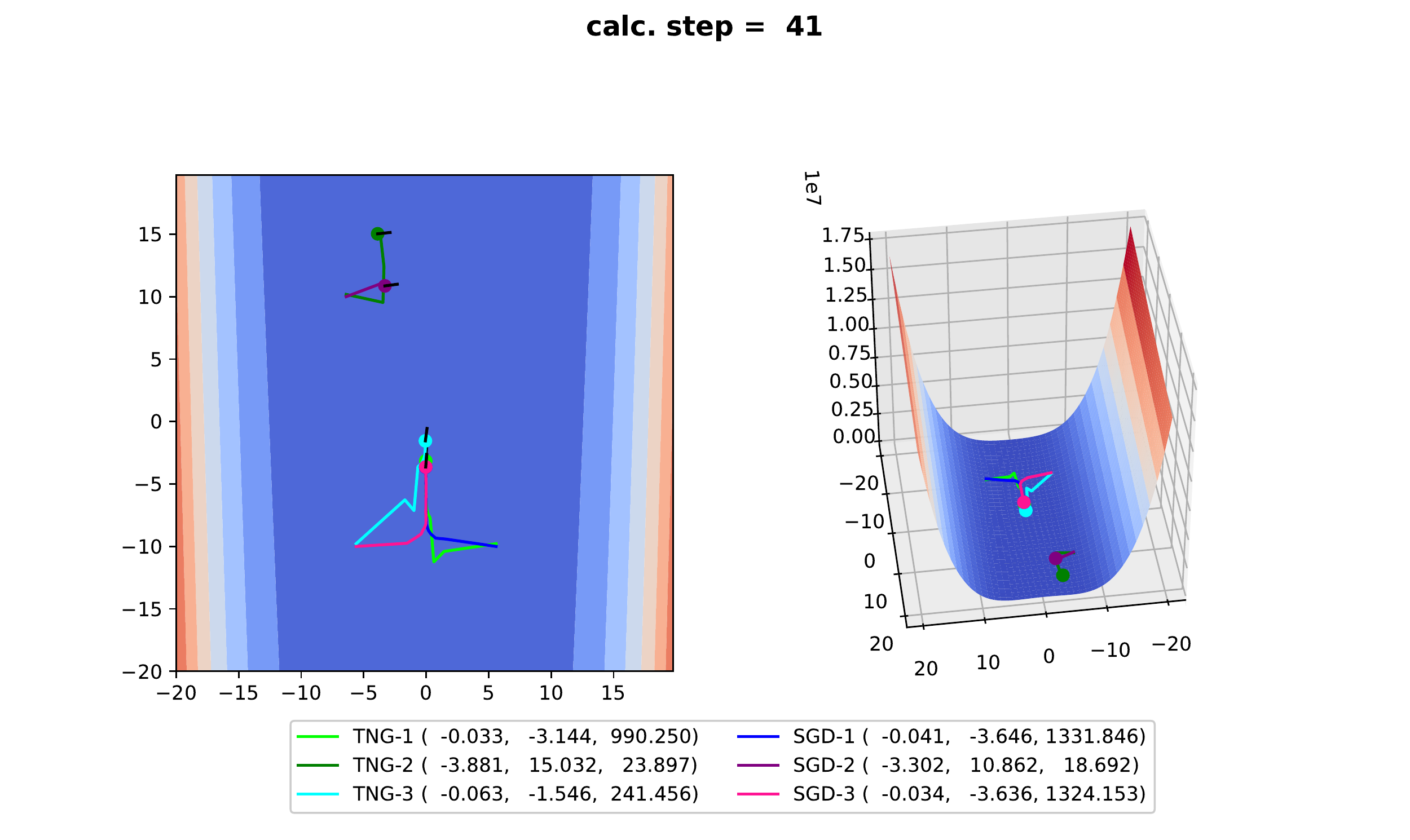}}
	\vspace{-10pt}
	\caption{ TNG on Benchmarking Nonconvex Functions. 
	}
	\vspace{-20pt}
	\label{fig:ak_plot}
\end{figure}

\begin{figure*}[!htbp]
\vspace{-7pt}
\centering
\subfigure 
{\includegraphics[width=0.95\textwidth]{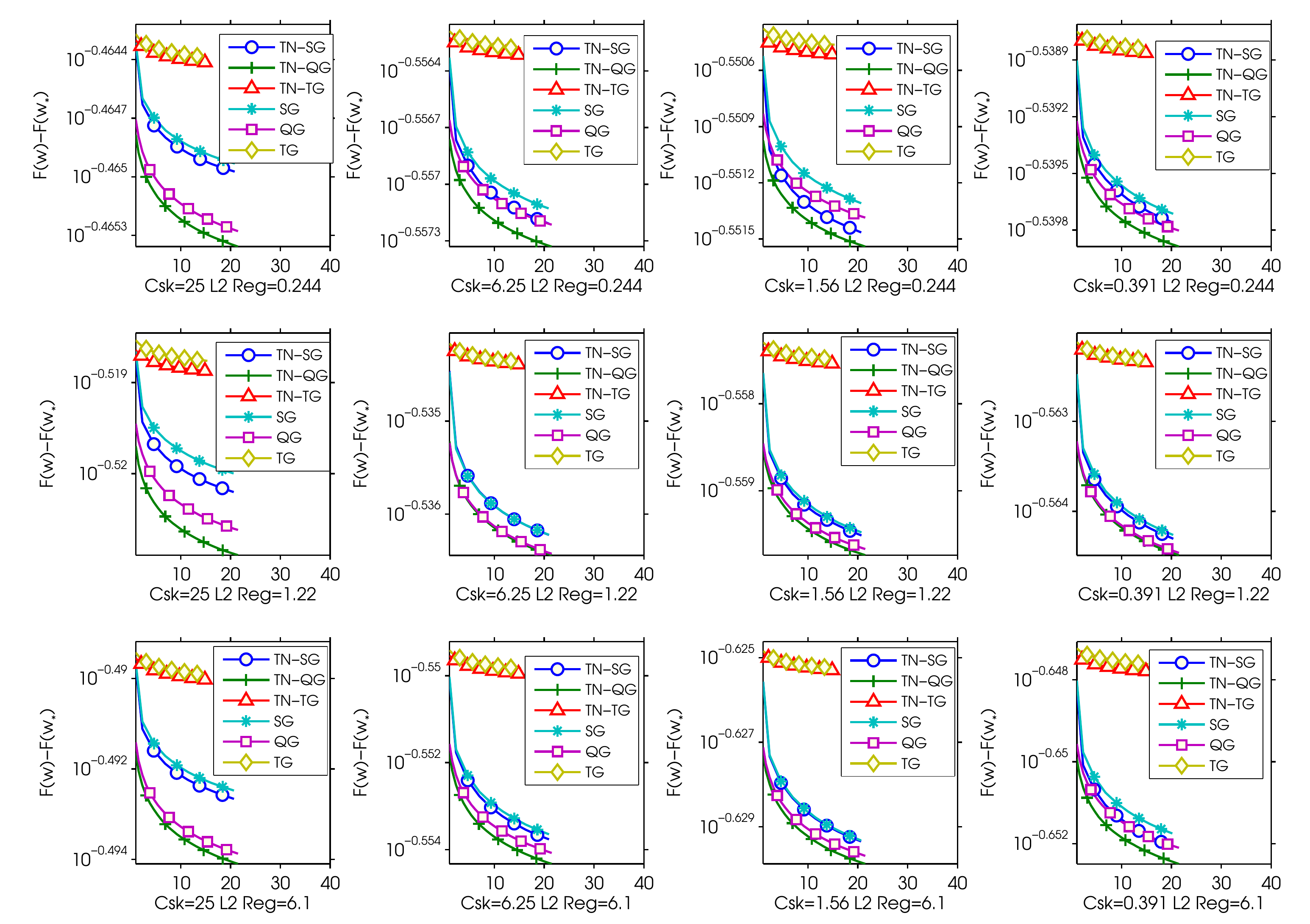}}
\vspace{-10pt}

\caption{ Convergence of SGD Methods. X-axis: (communications, bits for per element, Y-axis: the suboptimality $F(w_t)- F(\wmin)$.  
}
\vspace{-10pt}
\label{fig:cdgc1_com}
\end{figure*}

\begin{figure*}[h]
\vspace{-7pt}
\centering
\subfigure 
{\includegraphics[width=0.95\textwidth]{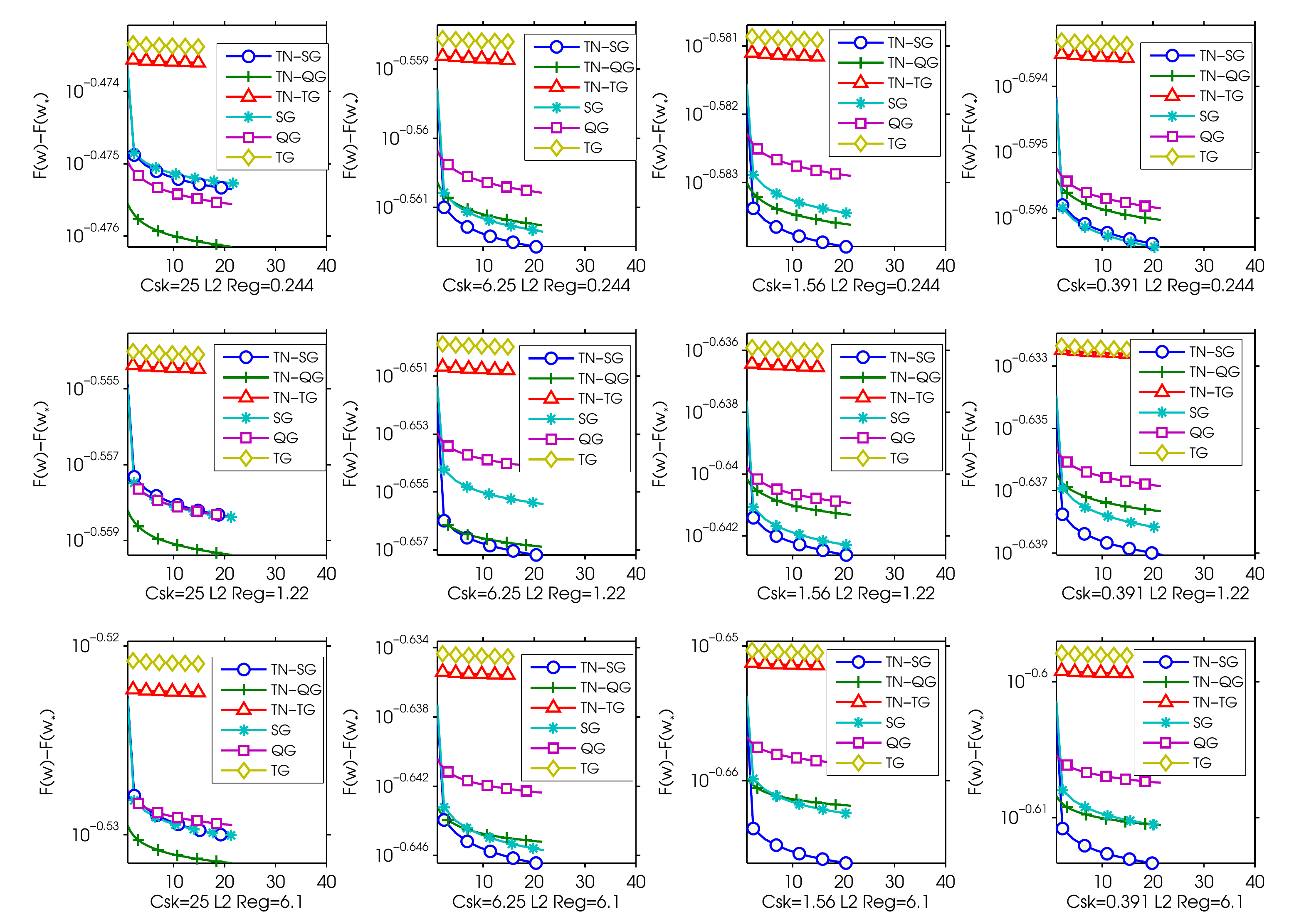}}
\vspace{-10pt}

\caption{ Convergence of Stochastic Quasi-Newton Methods. X-axis: (communications, bits for per element,Y-axis: the suboptimality. 
}
\vspace{-10pt}
\label{fig:cdgc2_com}
\end{figure*}

\begin{figure*}[h]
	\vspace{-7pt}
	\centering
	\subfigure 
	{\includegraphics[width=0.95\textwidth]{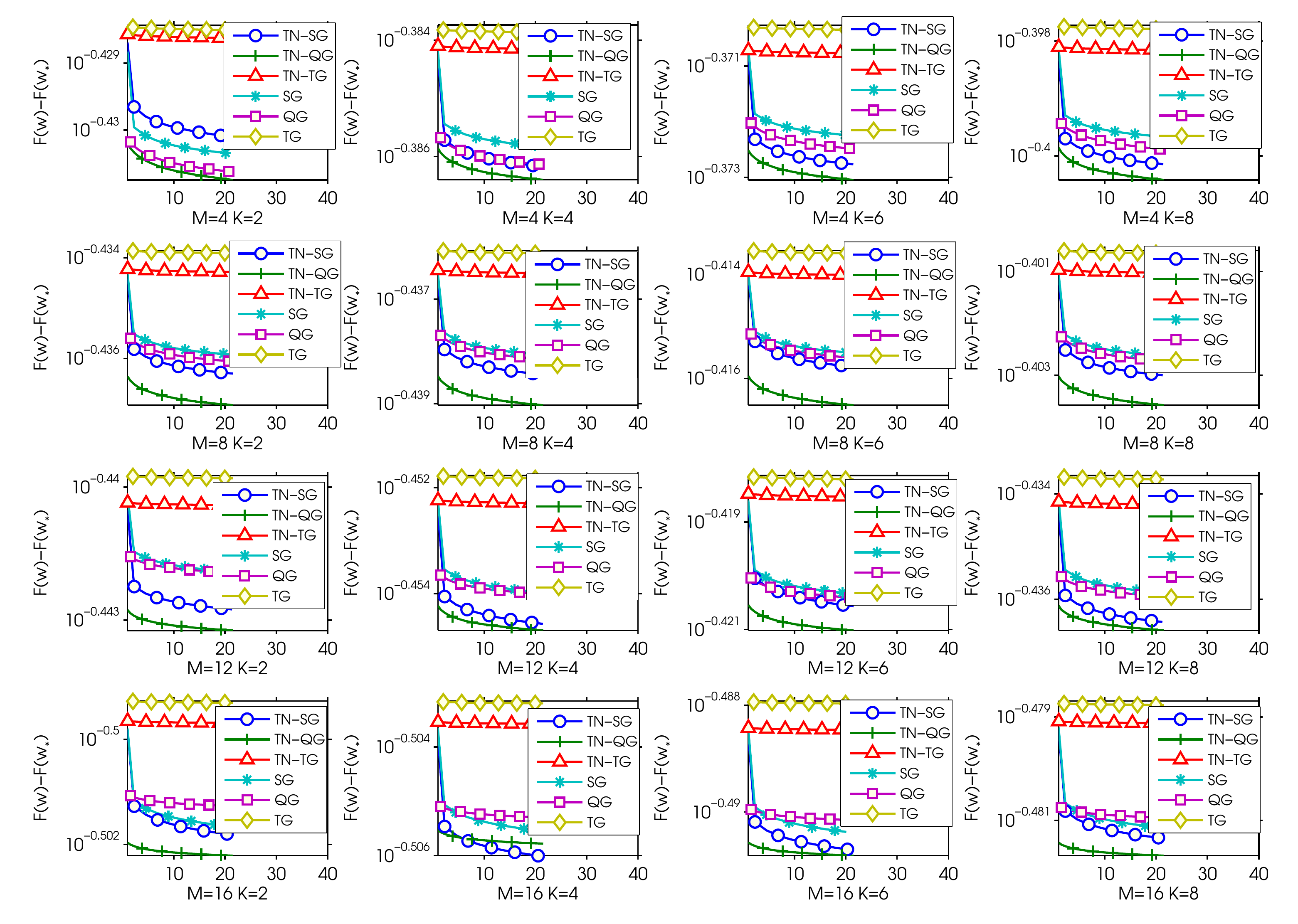}}
	
	\vspace{-10pt}

	\caption{ Convergence of Stochastic Quasi-Newton Methods. X-axis: (communications, bits for per element, Y-axis: the suboptimality. 
	}
	\vspace{-10pt}
	\label{fig:cdgc3_com}
\end{figure*}

\section{Experiments}

\subsection{Nonconvex Problems}
To visualize the efficiency of compressed normalized gradients on some hard non-convex functions, we plot some figures to demonstrate the optimization trajectories in  Figure \ref{fig:ak_plot}. These functions include  Ackley function ($f(x,y)=20 - 20 \exp(-0.2 \sqrt{0.5(x^2 + y^2)}) - \exp(0.5 (\cos(2 \pi x)+\sin(2 \pi y))) + e$, and global minimum at $f(0, 0) = 0$), Booth function ($(x + 2y - 7)^2 + (2 x + y - 5)^2$ and global minimum at $f(1,3) = 0$), and Rosenbrock function ($f(x,y)=100 (x - y^2)^2 + (x-1)^2$ and global minimum is at $ f(1,1) = 0$). The stochast gradient is synthetically generated by adding Gaussian noise, each element of which follows $\mathcal N(0,1)$, and step size is fixed through all iterations. We search for the optimal step size, and set $ \eta_t= 5\times 10^{-3}$ for Ackley function, $\eta_t= 10^{-4}$ for Booth function and $\eta_t= 10^{-6}$ for Rosenbrock function. 
Normalized gradients are noted in the figure as \textit{TNG} and the baseline noted as \textit{SGD}. We choose the ternary coding \cite{wen2017terngrad} of stochastic gradients for both methods, and the difference is with or without trajectory normalization.  For each optimizer, we noted the current parameter and objective function values as $(x,y,f(x,y))$ below each fiure. We make sure that two approaches use equal communication for a fair comparison, by counting one round of reference vector communication in \textit{16}-bits representation as $8$ iterations of pure ternary coding. The reference vector is chosen to be updated by every $16$ iterations. As non-convex optimization is sensitive to initialization points, we choose three initialization points, and we noted the optimizers with a number suffix to indicate different initializations. In general, the normalized gradient is compression-robust, as it converges faster. The improvement on the oscillating surface like Ackley function than flat surface like Rosenbrock, which aligns with our motivation that the compression error depends on the intrinsic distribution of gradients.

\subsection{Convex Problems}
We study the TNG combined with different kinds of gradients, coding strategies, reference gradient formulation, with or without second-order gradients, to prove the generality of the proposed methods. We use the mini-batch stochastic gradient descent, along with its quasi-Newton adaption \cite{byrd2016stochastic}. The stochastic quasi-Newton uses L-BFGS method for updating the Hessian matrix and stochastic gradients as the first-order gradient. To be specific, we replace the vanilla stochastic gradient with the second-order gradient $p_t = H_t g_t \in \mathbb R^D$, 
where  $H_t$ is an approximate inverse Hessian matrix by using the past trajectory of both parameters and gradients of within the memory (of size $K$)
\begin{eqnarray} 
s_k=w_k- w_{k-1}, \quad y_k=g_k- g_{k-1},\quad  \forall k  \in [t]
\end{eqnarray}
Denoting $\rho_t = 1/s_t^{\top} y_t$, we initialize it with $H_t^{t-K} = (s^{\top}_t y_t/|| y_t ||^2)I_D$,
where $I_D \in \mathbb R^{D \times D}$ is a diagonal matrix. Then L-BFGS udpates the inverse Hessian as
\begin{eqnarray}\label{eq:bfgs_update}
H_t^k=(I_D-\rho_k s_k y_k^{\top})^{\top} H_t^{k-1}(I-\rho_k s_k y_k^{\top})+\rho_k s_k s_k^{\top} 
\end{eqnarray}
for $t-K + 1 \leq k \leq t$, and finally generates $H_t \leftarrow H_t^t$.

We will mainly use the $\ell_2$-regularized logistic regression as a representative  convex problem to evaluate the efficiency. We use $ a_n \in \mathbb R^d$ representing a feature sample and $b_n \in \{-1,1\}$ represents its label. 
We use the same procedure with \cite{wangni2018gradient} to generate a large pool of synthetic data that have different scale of skewness of gradient distribution, with two hyperparameters $\csk$ and $\cskk$ that control the skewness: we sample normalized data vectors from standard Gaussian distribution for each element, 
\begin{align*}
\text{normalized data:} \quad &\bar a_{nd} \sim \mathcal N(0,1),\quad \forall d \in [D],n \in [N],
\end{align*}
meanwhile sample magnitude vectors from a uniform distribution, and the smaller magnitudes are shrunk so the distribution is skewed. The features are elementwise products of the two normalized data and the magnitudes individually. The data is $D=512$ dimensional and each setting generate a dataset of size $N=2048$.
\begin{align*}
\text{magnitudes:} \quad & \bar B \sim \text{Uniform}[0,1]^D,\quad \bar B_{d} \leftarrow \csk \bar B_{d}, \\
\text{if:}\quad &\bar B_d \leq \cskk,\quad \forall d \in [D] ,\quad a_{n} \leftarrow \bar a_{n} \odot \bar B, \\
\text{label:}  \quad &
\bar w \sim \mathcal N(0,I),\quad  b_n \leftarrow \sgn(\bar a_{n}^{\top}\bar w ).
\end{align*}
and a smaller $\csk$ implies a stronger skewness or sparsity in gradient distribution.
%


First, we simulated $M=4$ servers where the main server does the averaging and broadcasting jobs. We use two kinds of algorithms to calculate $g_t$: SGD and SVRG, the batch-size is always set to be $8$. We plotted the convergence behavior of them in Figure \ref{fig:cdgc1_com} respectively, in terms of communications, the product of the number of data passes and the compression rate of gradient information. We use $\cskk=0.6$ for all settings, and in the $i^{\text{th}}$ row and $j^{\text{th}}$ column, we set $\csk \propto 1/4^j$ and the $\ell_2$-regularization to $\lambda_2 \propto 1/2^i$ individually, to test the sensitivity of TNG under different level of convexity and gradient skewness. We compare our approach with gradient quantization \cite{alistarh2016qsgd} (noted as \textit{QG} in the figures), randomized ternary coding \cite{wen2017terngrad}  (noted as \textit{TG} in the figures) and gradient sparsification (noted as \textit{SG} in the figures) \cite{wangni2018gradient}, three approaches that favor different distributions for compression. We tuned the step-size for the fastest convergence speed, and found that under the general principle $\eta_t \propto 1/\textit{variance}$ perform stably for all methods, and a larger step-size caused divergence in some settings. We noticed that \textit{TG} methods have a larger variance than other two, therefore we measured their variance with a shrinking factor of $1/5$, to make it easy for plotting. In each subfiure, we noted the parameter $\csk$ and $\ell_2$ regularization (value $\times 100$ for showing). We also plotted in Figure \ref{fig:cdgc2_com} with convergence of the stochastic second-order gradient method, with exactly the same setting of convexity, sparsity, etc, with Figure \ref{fig:cdgc1_com}, respectively. We also test the sensitivity of settings like the number of servers and memory size of quasi-Newton methods, in Figure \ref{fig:cdgc3_com}. In the $i^{\text{th}}$ row and $j^{\text{th}}$ column, we set the number of servers to be $M=4 i$, and the memory size $K=2 j$, and the settings are noted below the subfigures.

Our normalization technique is combined with three kinds of codings, respectively. (noted with prefix \textit{TN} in the figure). We initialize the reference vector with a full gradient, and in the following iterations, the reference is updated to be the averaged compressed TNG $\gti=\sum_m v^m(w_{t-1})/M$ from the last iteration $t-1$. This can be done with a round of broadcasting for the reference vector in a synchronous setting, or the other servers can inference from the past parameters $\gti = (w_t-w_{t-1})/\eta$ without additional communication. The balance between the fitness of $\gti$ and its cost needs to be balanced for different problems. 
When calculating bits for each approach, we also choose the optimal methods for coding the vectors, whether in dense vector form or in sparse vector form, the latter of which suits a case where the distribution of $-1,0,1$ is uneven. 

By observing the figures, we see that the normalization clearly improves upon baselines, in basically all the settings, and the improvement gap has a dependence on conditions. 
Since difference coding strategies have advantages in different problems, we do not compare them with each other. The \textit{SG} methods, majorly use the bits for transmitting full-precision of important elements, and should be improved if using low-precision, i.e. quantized numbers. We found that TNG improves upon the baseline more under stronger convexity and weaker gradient skewness.  By comparing with different level of sparsity in gradient distribution, the different kinds of coding methods tend to have slightly different performance: for example, we see that  \textit{QG} is relatively insensitive to skewness of gradients comparing to  \textit{SG}, and  \textit{SG} performs better with stronger convexity. Besides, by observing Figure \ref{fig:cdgc3_com} vertically, a larger number of servers provides a better reference vector; and observing horizontallly, we see that increasing memory size initially improves convergence but gradually becomes ineffective.

\section{Related Works}
Researchers proposed protocols from other perspectives to reduce communication. 
A prevailing method is to average parameter occasionally, but not too frequent \cite{tsianos2012communication, wang2018cooperative}, or just one round of averaging over final parameters\cite{zhang2012communication}. If the problems require the servers to frequently synchronized, we can use an asynchronous protocol like parameter servers \cite{ho2013more,li2014communication}, where each server requests the latest parameter from the main server or contributes its gradients, passively or aggressively, based on the network condition; the decentralized optimization algorithms \cite{yuan2016convergence, lan2017communication, lian2017can} view every servers equally, to avoid the congestion of communication since the main server takes over most of the requests and causing unbalance.  Efficiently using a large batch-size \cite{cotter2011better, li2014efficient, wang2017improved, goyal2017accurate}  or the second-order gradient \citep{shamir2014communication, zhang2015disco} will reduce the communication since the overall number of iterations, and therefore reduce commnunication.
the model synchronization can also be formulated as a global consensus problem \cite{zhang2014asynchronous} with penalty of delay. 
Besides, the normalization idea was also used in other areas, like normalized gradient descent for general convex or quasi-convex optimization \cite{nesterov1984, hazan2015beyond}; on different subjects, normalization helps to stablize the feature or gradient distribution in neural networks  \cite{ioffe2015batch, klambauer2017self, neyshabur2015path, salimans2016weight}.



\section{Conclusion}
In this paper we propose a simple and general protocol, of using the trajectory normalized gradient, to reduce the compression error for gradient communication during distributed optimization. We provide insight to normalize gradient more accurately, and validate our idea on various experiments with different parameters and coding strategies. 



\nocite{langley00}
\newpage
\bibliography{icml19}

\begin{thebibliography}{45}
\providecommand{\natexlab}[1]{#1}
\providecommand{\url}[1]{\texttt{#1}}
\expandafter\ifx\csname urlstyle\endcsname\relax
  \providecommand{\doi}[1]{doi: #1}\else
  \providecommand{\doi}{doi: \begingroup \urlstyle{rm}\Url}\fi

\bibitem[Agarwal \& Duchi(2011)Agarwal and Duchi]{agarwal2011distributed}
Agarwal, A. and Duchi, J.~C.
\newblock Distributed delayed stochastic optimization.
\newblock In \emph{Advances in Neural Information Processing Systems}, pp.\
  873--881, 2011.

\bibitem[Aji \& Heafield(2017)Aji and Heafield]{aji2017sparse}
Aji, A.~F. and Heafield, K.
\newblock Sparse communication for distributed gradient descent.
\newblock In \emph{Proceedings of the 2017 Conference on Empirical Methods in
  Natural Language Processing}, pp.\  440--445, 2017.

\bibitem[Alistarh et~al.(2017)Alistarh, Grubic, Li, Tomioka, and
  Vojnovic]{alistarh2016qsgd}
Alistarh, D., Grubic, D., Li, J., Tomioka, R., and Vojnovic, M.
\newblock QSGD: Communication-efficient SGD via gradient quantization and
  encoding.
\newblock In \emph{Advances in Neural Information Processing Systems}, pp.\
  1707--1718, 2017.

\bibitem[Alistarh et~al.(2018)Alistarh, Hoefler, Johansson, Konstantinov,
  Khirirat, and Renggli]{alistarh2018convergence}
Alistarh, D., Hoefler, T., Johansson, M., Konstantinov, N., Khirirat, S., and
  Renggli, C.
\newblock The convergence of sparsified gradient methods.
\newblock In \emph{Advances in Neural Information Processing Systems}, pp.\
  5977--5987, 2018.

\bibitem[Bernstein et~al.(2018)Bernstein, Wang, Azizzadenesheli, and
  Anandkumar]{bernstein2018signsgd}
Bernstein, J., Wang, Y.-X., Azizzadenesheli, K., and Anandkumar, A.
\newblock SignSGD: compressed optimisation for non-convex problems.
\newblock \emph{arXiv preprint arXiv:1802.04434}, 2018.

\bibitem[Bottou(2010)]{bottou2010large}
Bottou, L.
\newblock Large-scale machine learning with stochastic gradient descent.
\newblock In \emph{Proceedings of COMPSTAT'2010}, pp.\  177--186. Springer,
  2010.

\bibitem[Bottou et~al.(2018)Bottou, Curtis, and
  Nocedal]{bottou2018optimization}
Bottou, L., Curtis, F.~E., and Nocedal, J.
\newblock Optimization methods for large-scale machine learning.
\newblock \emph{SIAM Review}, 60\penalty0 (2):\penalty0 223--311, 2018.

\bibitem[Byrd et~al.(2016)Byrd, Hansen, Nocedal, and
  Singer]{byrd2016stochastic}
Byrd, R.~H., Hansen, S.~L., Nocedal, J., and Singer, Y.
\newblock A stochastic quasi-Newton method for large-scale optimization.
\newblock \emph{SIAM Journal on Optimization}, 26\penalty0 (2):\penalty0
  1008--1031, 2016.

\bibitem[Cormen et~al.(2009)Cormen, Leiserson, Rivest, and
  Stein]{cormen2009introduction}
Cormen, T.~H., Leiserson, C.~E., Rivest, R.~L., and Stein, C.
\newblock \emph{Introduction to algorithms}.
\newblock MIT press, 2009.

\bibitem[Cotter et~al.(2011)Cotter, Shamir, Srebro, and
  Sridharan]{cotter2011better}
Cotter, A., Shamir, O., Srebro, N., and Sridharan, K.
\newblock Better mini-batch algorithms via accelerated gradient methods.
\newblock In \emph{Advances in Neural Information Processing Systems}, pp.\
  1647--1655, 2011.

\bibitem[Goyal et~al.(2017)Goyal, Doll{\'a}r, Girshick, Noordhuis, Wesolowski,
  Kyrola, Tulloch, Jia, and He]{goyal2017accurate}
Goyal, P., Doll{\'a}r, P., Girshick, R., Noordhuis, P., Wesolowski, L., Kyrola,
  A., Tulloch, A., Jia, Y., and He, K.
\newblock Accurate, large minibatch SGD: training imagenet in 1 hour.
\newblock \emph{arXiv preprint arXiv:1706.02677}, 2017.

\bibitem[Hazan et~al.(2015)Hazan, Levy, and Shalev-Shwartz]{hazan2015beyond}
Hazan, E., Levy, K., and Shalev-Shwartz, S.
\newblock Beyond convexity: Stochastic quasi-convex optimization.
\newblock In \emph{Advances in Neural Information Processing Systems}, pp.\
  1594--1602, 2015.

\bibitem[Ho et~al.(2013)Ho, Cipar, Cui, Lee, Kim, Gibbons, Gibson, Ganger, and
  Xing]{ho2013more}
Ho, Q., Cipar, J., Cui, H., Lee, S., Kim, J.~K., Gibbons, P.~B., Gibson, G.~A.,
  Ganger, G., and Xing, E.~P.
\newblock More effective distributed ml via a stale synchronous parallel
  parameter server.
\newblock In \emph{Advances in Neural Information Processing Systems}, pp.\
  1223--1231, 2013.

\bibitem[Ioffe \& Szegedy(2015)Ioffe and Szegedy]{ioffe2015batch}
Ioffe, S. and Szegedy, C.
\newblock Batch normalization: Accelerating deep network training by reducing
  internal covariate shift.
\newblock In \emph{International Conference on Machine Learning}, pp.\
  448--456, 2015.

\bibitem[Jin et~al.(2017)Jin, Ge, Netrapalli, Kakade, and
  Jordan]{jin2017escape}
Jin, C., Ge, R., Netrapalli, P., Kakade, S.~M., and Jordan, M.~I.
\newblock How to escape saddle points efficiently.
\newblock In \emph{International Conference on Machine Learning}, pp.\
  1724--1732, 2017.

\bibitem[Johnson \& Zhang(2013)Johnson and Zhang]{johnson2013accelerating}
Johnson, R. and Zhang, T.
\newblock Accelerating stochastic gradient descent using predictive variance
  reduction.
\newblock In \emph{Advances in Neural Information Processing Systems}, pp.\
  315--323, 2013.

\bibitem[Klambauer et~al.(2017)Klambauer, Unterthiner, Mayr, and
  Hochreiter]{klambauer2017self}
Klambauer, G., Unterthiner, T., Mayr, A., and Hochreiter, S.
\newblock Self-normalizing neural networks.
\newblock In \emph{Advances in Neural Information Processing Systems}, pp.\
  971--980, 2017.

\bibitem[Kleinberg et~al.(2018)Kleinberg, Li, and
  Yuan]{kleinberg2018alternative}
Kleinberg, R., Li, Y., and Yuan, Y.
\newblock An alternative view: When does SGD escape local minima?
\newblock \emph{arXiv preprint arXiv:1802.06175}, 2018.

\bibitem[Lan et~al.(2017)Lan, Lee, and Zhou]{lan2017communication}
Lan, G., Lee, S., and Zhou, Y.
\newblock Communication-efficient algorithms for decentralized and stochastic
  optimization.
\newblock \emph{arXiv preprint arXiv:1701.03961}, 2017.

\bibitem[Li et~al.(2014{\natexlab{a}})Li, Andersen, Smola, and
  Yu]{li2014communication}
Li, M., Andersen, D.~G., Smola, A.~J., and Yu, K.
\newblock Communication efficient distributed machine learning with the
  parameter server.
\newblock In \emph{Advances in Neural Information Processing Systems}, pp.\
  19--27, 2014{\natexlab{a}}.

\bibitem[Li et~al.(2014{\natexlab{b}})Li, Zhang, Chen, and
  Smola]{li2014efficient}
Li, M., Zhang, T., Chen, Y., and Smola, A.~J.
\newblock Efficient mini-batch training for stochastic optimization.
\newblock In \emph{Proceedings of the 20th ACM SIGKDD international conference
  on Knowledge discovery and data mining}, pp.\  661--670. ACM,
  2014{\natexlab{b}}.

\bibitem[Lian et~al.(2017)Lian, Zhang, Zhang, Hsieh, Zhang, and
  Liu]{lian2017can}
Lian, X., Zhang, C., Zhang, H., Hsieh, C.-J., Zhang, W., and Liu, J.
\newblock Can decentralized algorithms outperform centralized algorithms? a
  case study for decentralized parallel stochastic gradient descent.
\newblock In \emph{Advances in Neural Information Processing Systems}, pp.\
  5330--5340, 2017.

\bibitem[Nesterov(1984)]{nesterov1984}
Nesterov, Y.
\newblock Minimization methods for nonsmooth convex and quasiconvex functions.
\newblock \emph{Matekon}, pp.\  29:519–531, 1984.

\bibitem[Nesterov(2013)]{nesterov2013introductory}
Nesterov, Y.
\newblock \emph{Introductory lectures on convex optimization: A basic course},
  volume~87.
\newblock Springer Science \& Business Media, 2013.

\bibitem[Neyshabur et~al.(2015)Neyshabur, Salakhutdinov, and
  Srebro]{neyshabur2015path}
Neyshabur, B., Salakhutdinov, R.~R., and Srebro, N.
\newblock Path-SGD: Path-normalized optimization in deep neural networks.
\newblock In \emph{Advances in Neural Information Processing Systems}, pp.\
  2422--2430, 2015.

\bibitem[Nguyen et~al.(2018)Nguyen, Nguyen, van Dijk, Richt{\'a}rik,
  Scheinberg, and Tak{\'a}{\v{c}}]{nguyen2018sgd}
Nguyen, L.~M., Nguyen, P.~H., van Dijk, M., Richt{\'a}rik, P., Scheinberg, K.,
  and Tak{\'a}{\v{c}}, M.
\newblock SGD and Hogwild! convergence without the bounded gradients
  assumption.
\newblock \emph{arXiv preprint arXiv:1802.03801}, 2018.

\bibitem[Salimans \& Kingma(2016)Salimans and Kingma]{salimans2016weight}
Salimans, T. and Kingma, D.~P.
\newblock Weight normalization: A simple reparameterization to accelerate
  training of deep neural networks.
\newblock In \emph{Advances in Neural Information Processing Systems}, pp.\
  901--909, 2016.

\bibitem[Schmidt et~al.(2017)Schmidt, Le~Roux, and Bach]{schmidt2017minimizing}
Schmidt, M., Le~Roux, N., and Bach, F.
\newblock Minimizing finite sums with the stochastic average gradient.
\newblock \emph{Mathematical Programming: Series A and B}, 162\penalty0
  (1-2):\penalty0 83--112, 2017.

\bibitem[Shamir et~al.(2014)Shamir, Srebro, and Zhang]{shamir2014communication}
Shamir, O., Srebro, N., and Zhang, T.
\newblock Communication-efficient distributed optimization using an approximate
  newton-type method.
\newblock In \emph{International Conference on Machine Learning}, pp.\
  1000--1008, 2014.

\bibitem[Stich et~al.(2018)Stich, Cordonnier, and Jaggi]{stich2018sparsified}
Stich, S.~U., Cordonnier, J.-B., and Jaggi, M.
\newblock Sparsified SGD with memory.
\newblock In \emph{Advances in Neural Information Processing Systems}, pp.\
  4452--4463, 2018.

\bibitem[Szeliski(2010)]{szeliski2010computer}
Szeliski, R.
\newblock \emph{Computer vision: algorithms and applications}.
\newblock Springer Science \& Business Media, 2010.

\bibitem[Tsianos et~al.(2012)Tsianos, Lawlor, and
  Rabbat]{tsianos2012communication}
Tsianos, K., Lawlor, S., and Rabbat, M.~G.
\newblock Communication/computation tradeoffs in consensus-based distributed
  optimization.
\newblock In \emph{Advances in Neural Information Processing Systems}, pp.\
  1943--1951, 2012.

\bibitem[Wang et~al.(2018)Wang, Sievert, Liu, Charles, Papailiopoulos, and
  Wright]{wang2018atomo}
Wang, H., Sievert, S., Liu, S., Charles, Z., Papailiopoulos, D., and Wright, S.
\newblock Atomo: Communication-efficient learning via atomic sparsification.
\newblock In \emph{Advances in Neural Information Processing Systems}, pp.\
  9872--9883, 2018.

\bibitem[Wang \& Joshi(2018)Wang and Joshi]{wang2018cooperative}
Wang, J. and Joshi, G.
\newblock Cooperative SGD: A unified framework for the design and analysis of
  communication-efficient SGD algorithms.
\newblock \emph{arXiv preprint arXiv:1808.07576}, 2018.

\bibitem[Wang \& Zhang(2017)Wang and Zhang]{wang2017improved}
Wang, J. and Zhang, T.
\newblock Improved optimization of finite sums with minibatch stochastic
  variance reduced proximal iterations.
\newblock \emph{arXiv preprint arXiv:1706.07001}, 2017.

\bibitem[Wangni et~al.(2018)Wangni, Wang, Liu, and Zhang]{wangni2018gradient}
Wangni, J., Wang, J., Liu, J., and Zhang, T.
\newblock Gradient sparsification for communication-efficient distributed
  optimization.
\newblock In \emph{Advances in Neural Information Processing Systems}, pp.\
  1306--1316, 2018.

\bibitem[Wen et~al.(2017)Wen, Xu, Yan, Wu, Wang, Chen, and Li]{wen2017terngrad}
Wen, W., Xu, C., Yan, F., Wu, C., Wang, Y., Chen, Y., and Li, H.
\newblock Terngrad: Ternary gradients to reduce communication in distributed
  deep learning.
\newblock \emph{arXiv preprint arXiv:1705.07878}, 2017.

\bibitem[Wright \& Nocedal(1999)Wright and Nocedal]{wright1999numerical}
Wright, S. and Nocedal, J.
\newblock Numerical optimization.
\newblock \emph{Springer Science}, 35\penalty0 (67-68):\penalty0 7, 1999.

\bibitem[Wu et~al.(2018)Wu, Huang, Huang, and Zhang]{wu2018error}
Wu, J., Huang, W., Huang, J., and Zhang, T.
\newblock Error compensated quantized SGD and its applications to large-scale
  distributed optimization.
\newblock \emph{arXiv preprint arXiv:1806.08054}, 2018.

\bibitem[Yuan et~al.(2016)Yuan, Ling, and Yin]{yuan2016convergence}
Yuan, K., Ling, Q., and Yin, W.
\newblock On the convergence of decentralized gradient descent.
\newblock \emph{SIAM Journal on Optimization}, 26\penalty0 (3):\penalty0
  1835--1854, 2016.

\bibitem[Zhang \& Kwok(2014)Zhang and Kwok]{zhang2014asynchronous}
Zhang, R. and Kwok, J.
\newblock Asynchronous distributed admm for consensus optimization.
\newblock In \emph{International Conference on Machine Learning}, pp.\
  1701--1709, 2014.

\bibitem[Zhang(2004)]{zhang2004solving}
Zhang, T.
\newblock Solving large scale linear prediction problems using stochastic
  gradient descent algorithms.
\newblock In \emph{Proceedings of the twenty-first International Conference on
  Machine Learning}, pp.\  116. ACM, 2004.

\bibitem[Zhang \& Lin(2015)Zhang and Lin]{zhang2015disco}
Zhang, Y. and Lin, X.
\newblock Disco: Distributed optimization for self-concordant empirical loss.
\newblock In \emph{International Conference on Machine Learning}, pp.\
  362--370, 2015.

\bibitem[Zhang et~al.(2012)Zhang, Wainwright, and
  Duchi]{zhang2012communication}
Zhang, Y., Wainwright, M.~J., and Duchi, J.~C.
\newblock Communication-efficient algorithms for statistical optimization.
\newblock In \emph{Advances in Neural Information Processing Systems}, pp.\
  1502--1510, 2012.

\bibitem[Zhou et~al.(2016)Zhou, Wu, Ni, Zhou, Wen, and Zou]{zhou2016dorefa}
Zhou, S., Wu, Y., Ni, Z., Zhou, X., Wen, H., and Zou, Y.
\newblock Dorefa-net: Training low bitwidth convolutional neural networks with
  low bitwidth gradients.
\newblock \emph{arXiv preprint arXiv:1606.06160}, 2016.

\end{thebibliography}
\bibliographystyle{icml2019}

\end{document}